\documentclass[
]{ceurart}

\usepackage{listings}
\usepackage{booktabs}
\usepackage{hyperref}
\usepackage{adjustbox}
\usepackage{subcaption}
\usepackage{url}
\begin{document}

\copyrightyear{2024}
\copyrightclause{Copyright for this paper by its authors.
  Use permitted under Creative Commons License Attribution 4.0
  International (CC BY 4.0).}

\conference{PhysioCHI: Towards Best Practices for Integrating Physiological Signals in HCI,   May 11, 2024, Honolulu, HI, USA}

\title{Camera-Based Remote Physiology Sensing for Hundreds of Subjects Across Skin Tones}

% %%
% %% The "author" command and its associated commands are used to define
% %% the authors and their affiliations.
% \author[1,2]{Dmitry S. Kulyabov}[%
% orcid=0000-0002-0877-7063,
% email=kulyabov-ds@rudn.ru,
% url=https://yamadharma.github.io/,
% ]
% \cormark[1]
% \address[1]{Peoples' Friendship University of Russia (RUDN University),
%   6 Miklukho-Maklaya St, Moscow, 117198, Russian Federation}
% \address[2]{Joint Institute for Nuclear Research,
%   6 Joliot-Curie, Dubna, Moscow region, 141980, Russian Federation}

% \author[3]{Ilaria Tiddi}[%
% orcid=0000-0001-7116-9338,
% email=i.tiddi@vu.nl,
% url=https://kmitd.github.io/ilaria/,
% \address[3]{Vrije Universiteit Amsterdam, De Boelelaan 1105, 1081 HV Amsterdam, The Netherlands}

% \author[4]{Manfred Jeusfeld}[%
% orcid=0000-0002-9421-8566,
% email=Manfred.Jeusfeld@acm.org,
% url=http://conceptbase.sourceforge.net/mjf/,
% \address[4]{University of Skövde, Högskolevägen 1, 541 28 Skövde, Sweden}
\author[1]{Jiankai Tang}
\fnmark[1]
\address{Tsinghua University, Beijing, China}
\author[1]{Xinyi Li}
\fnmark[1]
\author[1]{Jiacheng Liu}
\author[1]{Xiyuxing Zhang}
\author[1]{Zeyu Wang}
\author[1]{Yuntao Wang}
\cormark[1]
% %% Footnotes
\fntext[1]{These authors contributed equally.}
\cortext[1]{Corresponding author.}

\vspace{-0.5cm}
\begin{abstract}
Remote photoplethysmography (rPPG) emerges as a promising method for non-invasive, convenient measurement of vital signs, utilizing the widespread presence of cameras. Despite advancements, existing datasets fall short in terms of size and diversity, limiting comprehensive evaluation under diverse conditions. This paper presents an in-depth analysis of the VitalVideo dataset~\cite{toye2023vital}, the largest real-world rPPG dataset to date, encompassing 893 subjects and 6 Fitzpatrick skin tones. Our experimentation with six unsupervised methods and three supervised models demonstrates that datasets comprising a few hundred subjects(i.e., 300 for UBFC-rPPG, 500 for PURE, and 700 for MMPD-Simple) are sufficient for effective rPPG model training. Our findings highlight the importance of diversity and consistency in skin tones for precise performance evaluation across different datasets.
\end{abstract}


\maketitle
\vspace{-0.5cm}
\section{Introduction}

Convenient and accurate access to vital signs is crucial for assessing health states and preventing diseases. Traditional methods for measuring physiological parameters, such as pulse waves, heart rate, and blood oxygen levels, often rely on wearable devices. While effective, these methods pose risks of cross-infection and may not be suitable for infants due to their delicate fingers. In contrast, camera-based physiological sensing offers a novel, contactless, and non-invasive approach that is both rapid and accessible, leveraging the ubiquitous nature of cameras to transform health signal measurement into a more user-friendly experience for everybody. This technology has the potential to be applied in various scenarios, including remote patient monitoring~\cite{yan2020high}, fitness and wellness tracking~\cite{tang2023mmpd}, and in situations where traditional wearables are impractical or undesired~\cite{villarroel2019non}.

\begin{figure}[htp]
    \centering
    \includegraphics[width=1\linewidth]{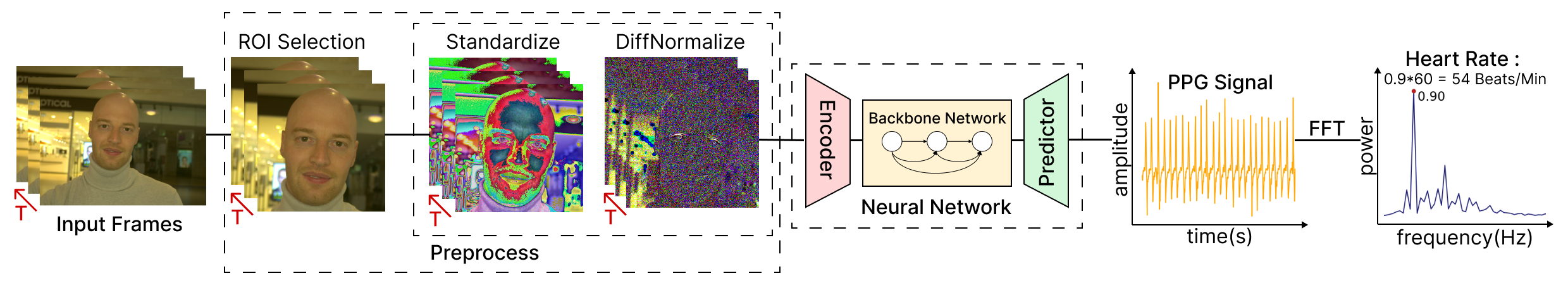}
    \caption{\textbf{Standard procedure.} Pipeline for predicting rPPG waveforms from facial frames, utilizing configuration settings consistent with the rPPG-Toolbox~\cite{liu2022rppg}}
    \label{fig: procedure}
    \vspace{-0.5cm}
\end{figure}
The underlying principle of camera-based physiological sensing, particularly remote photoplethysmography (rPPG), utilizes pixel information from video recordings to detect changes in light modulated by blood flow beneath the skin. These changes, reflecting the volumetric variations of oxygenated and deoxygenated hemoglobin, enable the measurement of the peripheral blood volume pulse (BVP), with rPPG signals closely correlating with heart rate~\cite{wang2016algorithmic,xu2014robust}. Leveraging advanced computer vision and time series analysis techniques, rPPG signal extraction emerges as a viable path for non-invasive health monitoring.  The standard process for extracting rPPG signals from facial footage encompasses several critical steps~\cite{liu2022rppg}: selecting the region of interest (ROI), performing signal normalization, training the neural network,  conducting model inference, and signal processing, as depicted in Fig \ref{fig: procedure}.

Despite the potential, the application of rPPG in real-world scenarios faces challenges, including body motion, varying lighting conditions, and poor-quality recordings. Moreover, accurately extracting pulse signals from individuals with darker skin tones remains difficult due to their lower light absorption and the risk of overexposure~\cite{tang2023mmpd,takano2007heart, nowara2020meta}. These challenges highlight the need for extensive research across diverse subjects and conditions. The emergence of the VitalVideo dataset~\cite{toye2023vital} presents an opportunity to explore the boundaries of rPPG technology with respect to subject size and skin tone diversity. This dataset offers a solid foundation for research, yet, intriguingly, it has not been utilized for comprehensive benchmarking or classification studies, including by its own creators. The availability of the VitalVideo dataset opens new avenues for addressing the aforementioned challenges, promising advancements in the accuracy and applicability of rPPG technologies across diverse populations and environments.

This paper enhances the understanding of camera-based health monitoring by making key contributions in a clear and straightforward manner:
1) We are the first to train and test on the VitalVideo dataset, sharing our code openly\footnote{\url{https://github.com/Health-HCI-Group/Largest_rPPG_Dataset_Evaluation}} and confirming the dataset's reliability for research.
2) By employing four datasets that include around 1000 individuals across 6 skin tones, we enhance benchmark efforts and underscore the importance of diverse data for model training.
3) Our analysis shows that optimal rPPG model training can be achieved with data from just a few hundred individuals. Additionally, we find that models are more sensitive to consistency in darker skin tones, highlighting the impact of data volume and skin tone diversity on model performance.

% contribution summarize
\vspace{-0.3cm}
\section{Related Works}
% Research About BenchMark（subject num， skin tone）  
\subsection{Dataset}
\begin{figure}[htp]
    \centering
    \includegraphics[width=0.8\linewidth]{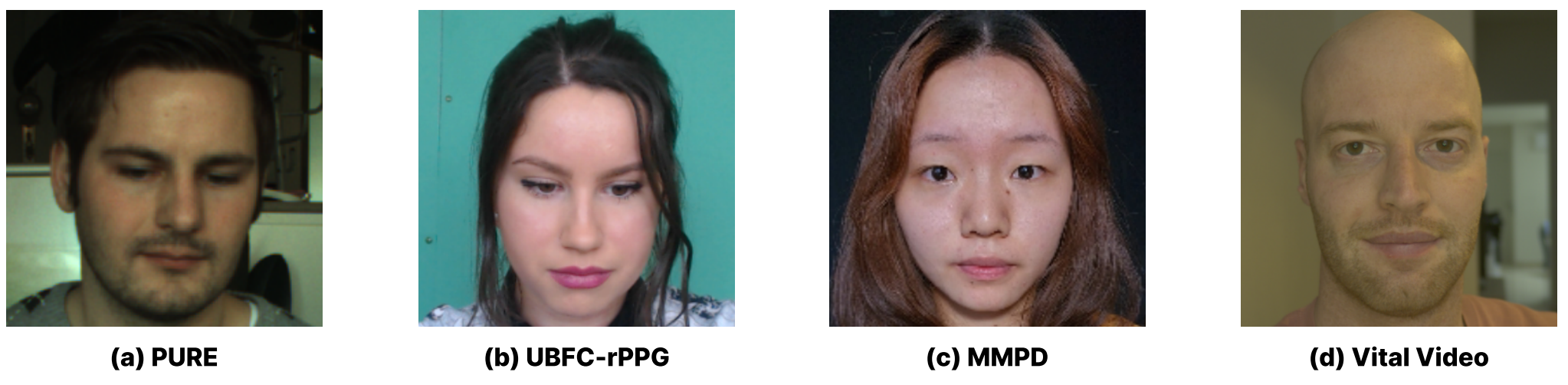}
    \vspace{-0.2cm}
    \caption{\textbf{Datasets Samples, including  PURE~\cite{stricker2014non},UBFC-rPPG~\cite{bobbia2019unsupervised},
    MMPD~\cite{tang2023mmpd}, VitalVideo~\cite{toye2023vital}}}
    \label{fig: Dataset}
    \vspace{-0.5cm}
\end{figure}
\vspace{-0.3cm}
From the multitude of available rPPG datasets, we carefully chose four representative datasets for our data training and testing, as depicted in Fig \ref{fig: Dataset}. The PURE~\cite{stricker2014non} and UBFC~\cite{bobbia2019unsupervised} datasets stand out as the most commonly used in rPPG research due to their reliability. MMPD~\cite{tang2023mmpd} distinguishes itself as the largest rPPG dataset captured using mobile phones, whereas VitalVideo~\cite{toye2023vital} boasts the greatest number of subjects and all Fitzpatrick skin tones.

\textbf{PURE}~\cite{stricker2014non}: Captured using an RGB eco274CVGE camera from SVS-Vistek GmbH, the dataset comprises recordings from 10 subjects, with a distribution of 8 males and 2 females. Video footage, obtained at a frequency of 30Hz and a resolution of 640x480, presents subjects positioned around 1.1 meters away from the camera. Illumination is provided by natural light filtering through a window, ensuring front-facing illumination. Gold-standard PPG and SpO2 data, sampled at 60Hz, are acquired through a CMS50E pulse oximeter affixed to the subject's finger. Across six recordings per participant, variations in motion conditions yield a diverse dataset reflecting different physiological states.

\textbf{UBFC-rPPG}~\cite{bobbia2019unsupervised}: Recorded using a Logitech C920 HD Pro webcam at 30Hz, the dataset features RGB videos with a resolution of 640x480, stored in an uncompressed 8-bit RGB format. Reference PPG data, serving as the gold-standard validation, is collected via a CMS50E transmissive pulse oximeter. Subjects, positioned approximately one meter from the camera, are recorded under indoor conditions illuminated by a combination of natural sunlight and artificial light sources.

\textbf{MMPD}~\cite{tang2023mmpd}: Comprising 660 one-minute videos captured using a Samsung Galaxy S22 Ultra mobile phone, the dataset is recorded at 30 frames per second with a resolution initially at 1280x720 pixels and later compressed to 320x240 pixels. Ground truth PPG signals, obtained simultaneously at 200 Hz using an HKG-07C+ oximeter and downsampled to 30 Hz, accompany the videos. The dataset encompasses Fitzpatrick skin types 3-6, four lighting conditions (LED-low, LED-high, incandescent, natural), and diverse activities including stationary, head rotation, talking, walking, and exercise scenarios.

\textbf{VitalVideo}~\cite{toye2023vital}: Utilizing a high-resolution Basler acA1920-40uc camera paired with a Basler C11-1220-12M lens, this dataset captures videos at 1920x1200 pixels, 30 fps. Participants, positioned 50 to 70 cm away, were recorded under varied indoor lighting, including artificial and natural sources, ensuring a minimum illumination of 150 lux with additional ring light support when needed. Accompanying physiological data were collected using a Contec CMS50D+ pulse oximeter and an Omron M7 blood pressure monitor, offering comprehensive multimodal data. Video clips, 30 seconds in length, recorded twice for each subject, were encoded in MP4 format, each approximately 2GB. The dataset, enriched with PPG waveforms, heart rate, SpO2, and blood pressure measurements, spans multiple indoor environments from libraries to shopping malls across ten locations in Western Europe, encompassing a diverse range of six skin tones and age groups ranging from 5 to 95 years. VV100, a subset with 100 subjects showcasing maximal demographic diversity, and the entire VVAll set with around 900 subjects were utilized for analysis.
% \textbf{VitalVideo}~\cite{toye2023vital}: The dataset was meticulously collected using a Basler acA1920-40uc camera equipped with a Basler C11-1220-12M lens, establishing a high-resolution video capture at 1920x1200 pixels and 30 frames per second. Participants, after giving consent, were positioned 50 to 70 cm in front of the camera mounted on a tripod. In addition to video data, physiological measurements were obtained via a Contec CMS50D+ pulse oximeter, with PPG and heart rate/SpO2 values sampled at 60Hz and 1Hz respectively, and a blood pressure monitor, the Omron M7. The unique setup facilitated the simultaneous recording of 30-second video clips alongside PPG waveforms, heart rate, SpO2, and blood pressure measurements, enhancing the dataset with rich multimodal data. Videos were encoded using libx264 codec in an MP4 container, each with an approximate file size of 2GB for 30 seconds of footage. To accommodate diverse lighting conditions across various indoor environments—ranging from libraries to shopping malls—recordings were made under both artificial and natural light, supplemented with a ring light when necessary to ensure a minimum illumination of 150 lux. This careful attention to detail in data collection and the environment ensures a dataset that is not only comprehensive but also reflective of a variety of lighting and background settings. To be noted, we utilized the subset VV100(100 subjects with maximal demographic diversity) and full set VVALL as catagried in original paper~\cite{toye2023vital}.
%VV100,VVAll,MMPD,PURE,UBFC-rPPG
\vspace{-0.2cm}
\subsection{Existing Methods}
\vspace{-0.1cm}
Recent advances in computer vision have significantly propelled the development of camera-based physiological vital sign measurements, introducing an expanding realm characterized by both unsupervised and supervised methodologies~\cite{mcduff2021camera}. Unsupervised methods utilize traditional signal processing and linear algebra to deduce PPG signals from RGB videos through techniques such as exploiting the green channel information (Green~\cite{verkruysse2008remote}), Independent Component Analysis (ICA~\cite{poh2010advancements}), chrominance-based signals (CHROM~\cite{de2013robust}), the plane orthogonal to skin (POS~\cite{wang2016algorithmic}), blood volume pulse extraction (PBV~\cite{de2014improved}), and motion-invariant features analysis (LGI~\cite{pilz2018local}). On the other hand, supervised methodologies employ neural networks, including DeepPhys~\cite{chen_deepphys_2018}, PhysNet~\cite{yu2019remote}, TS-CAN~\cite{liu2020multi}, and video transformer-based PhysFormer~\cite{yu2022physformer}, to improve accuracy through deep learning techniques and optimized training protocols. These advancements have been further bolstered by innovative data generation~\cite{mcduff2020advancing}, parallel computing~\cite{liu2024spikingphysformer} and augmentation strategies~\cite{wang2022synthetic}, as well as meta-learning~\cite{10.1145/3517225,liu2021metaphys}, federated learning~\cite{liu2022federated}, and unsupervised pretraining~\cite{sun2022contrast,gideon2021way,speth2023non,yang2022simper}, significantly elevating the state-of-the-art in non-invasive physiological measurement.
\vspace{-0.2cm}

\subsection{Benchmarks}
\vspace{-0.1cm}
Current benchmarks in the rPPG field concentrate on equitable evaluations across datasets for the task of heart rate estimation, comparing various networks and models~\cite{liu2022rppg,wang2023physbench}. The rPPG-Toolbox includes six unsupervised methods and six supervised networks for datasets such as PURE~\cite{stricker2014non}, UBFC-rPPG~\cite{bobbia2019unsupervised}, and MMPD~\cite{tang2023mmpd}. Nevertheless, there is a scarcity of research dedicated to analyzing each scenario in-depth to comprehend its limitations and optimal performance. MMPD~\cite{tang2023mmpd} stands out by offering eight labels for detailed examination, yet it lacks data for Fitzpatrick skin tone types 1 and 2. We have incorporated a data loader and label filter in the toolbox to facilitate the selection of all subset combinations (i.e., gender, skin tone, location, movements, and talking) for the VitalVideo dataset~\cite{toye2023vital}. To the best of our knowledge, our study is the first to conduct research on the largest VitalVideo dataset~\cite{toye2023vital} and examine skin tone label combinations across VitalVideo~\cite{toye2023vital} and MMPD~\cite{tang2023mmpd}.
\vspace{-0.5cm}
\section{Experiments}
\vspace{-0.1cm}
\subsection{Setup} 
% Enviroment, Data choose
%VV100,VVAll,MMPD,PURE,UBFC-rPPG,usage
For our experiments, we utilized an NVIDIA GeForce RTX 3090 graphics card, CUDA Version 12.2, and PyTorch version 2.1.1. We integrated the VitalVideo~\cite{toye2023vital} dataset with the rPPG-Toolbox~\cite{liu2022rppg}, introducing a label selector specifically for VitalVideo~\cite{toye2023vital} to streamline data extraction. The implemented code has been open-sourced at GitHub: \url{https://github.com/Health-HCI-Group/Largest_rPPG_Dataset_Evaluation}. The dataset was divided into VVAll~\cite{toye2023vital}(893 subjects), encompassing all data, and VV100~\cite{toye2023vital}(100 subjects), a more balanced subset. The VitalVideo~\cite{toye2023vital} dataset labels skin tones from type 1 to 6 according to Fitzpatrick rules. We selected MMPD's simplest subset(MMPD-Simple)~\cite{tang2023mmpd} by excluding natural lighting and motion conditions while preserving skin color variations. We chose settings like "LED-Low," "LED-High," and "Incandescent" for lighting, "Stationary" for motion, "False" for exercise, covering all skin tone types.

This study primarily focuses on cross-dataset training and testing, utilizing the VitalVideo~\cite{toye2023vital} dataset in conjunction with three other datasets: PURE~\cite{stricker2014non}, UBFC~\cite{bobbia2019unsupervised}, and MMPD~\cite{tang2023mmpd}. Our methodology commenced with applying six unsupervised methods~\cite{liu2022rppg} on the VitalVideo~\cite{toye2023vital} dataset. Subsequently, we conducted training sessions on various skin tones within the VV100~\cite{toye2023vital} and executed tests on the PURE~\cite{stricker2014non}, UBFC~\cite{bobbia2019unsupervised}, and MMPD~\cite{tang2023mmpd} datasets. Meanwhile, training was performed on the PURE~\cite{stricker2014non}, UBFC~\cite{bobbia2019unsupervised}, and MMPD~\cite{tang2023mmpd} datasets, followed by testing on each distinct skin tone category within VV100~\cite{toye2023vital}.

In our research, we selected six widely utilized unsupervised methods: ICA~\cite{poh2010advancements}, POS~\cite{wang2016algorithmic}, CHROM~\cite{de2013robust}, GREEN~\cite{verkruysse2008remote}, LGI~\cite{pilz2018local}, and PBV~\cite{de_Haan_2014}. These methods represent a broad spectrum of approaches in the field of rPPG technology, each with its unique advantages and applications. Additionally, we employed three state-of-the-art networks that are representative of the most advanced architectures in the domain: TS-CAN~\cite{liu2020multi}, which exemplifies 2D CNNs; PhysNet~\cite{yu2019remote}, a representative of 3D CNNs; and PhysFormer~\cite{yu2022physformer}, which utilizes the Transformer architecture. These selections were made to encompass a comprehensive range of methodologies, from traditional unsupervised techniques to cutting-edge neural networks, ensuring a thorough investigation into the capabilities and performance of camera-based physiological measurement technologies.
\vspace{-0.2cm}
\subsection{Unsupervised methods on VV100 and VVAll~\cite{toye2023vital}} %full and skin tone
% six methods for all and single 6 skin tones

\vspace{-0.2cm}
\begin{figure}[htp]
    \centering
    \includegraphics[width=0.9\linewidth]{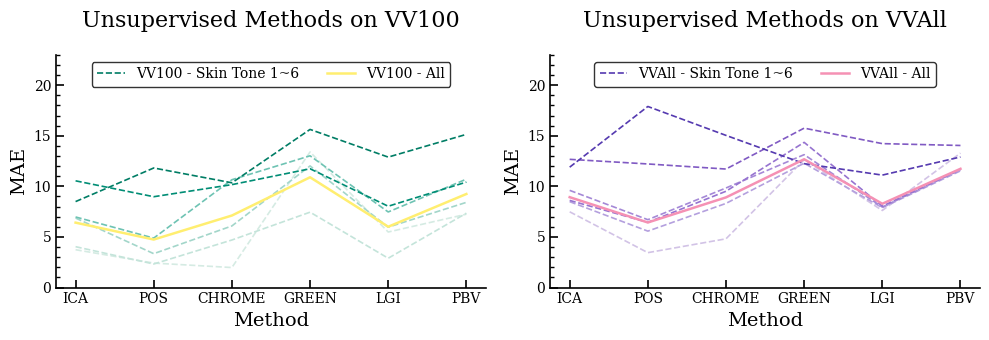}
    \caption{\textbf{Results of Unsupervised Methods on VitalVideo~\cite{toye2023vital}.} Line colors get deeper as skin tone varies from type 1 to type 6. MAE = Mean Absolute Error in HR estimation (Beats/Min). }
    \label{fig: Unsupervised}
\vspace{-0.4cm}   
\end{figure}

As demonstrated in Fig \ref{fig: Unsupervised}, the mean absolute errors (MAE) for both VV100 and VVAll~\cite{toye2023vital} exhibit an ascending trend as the skin tones range from type 1 to type 6 with POS~\cite{wang2016algorithmic} demonstrating the best performance in most cases. The similar trend observed in both subgraphs underscores the ability of the VV100 subset to effectively represent the larger VitalVideo dataset~\cite{toye2023vital}, demonstrating its comprehensiveness and relevance. Moreover, VV100~\cite{toye2023vital} not only mirrors the characteristics of VVAll~\cite{toye2023vital} but also has a more balanced distribution(e.g., skin tone),  making it a suitable subset for subsequent experiments. 

%分布一致则用vv100作为后续的标准数据集
\vspace{-0.1cm}
\subsection{Supervised methods trained on VV100~\cite{toye2023vital}} %Train： vv100 all and single 6 skin tones：7
% Test： MMPD（Simplest Subset）,PURE,UBFC-rPPG：3
%TSCAN（2D+TSM） Physnet（3D） PhysFormer_small（Transformer）：3

\begin{table*}[h!]
    \small
	\caption{\textbf{Supervised methods trained on VV100~\cite{toye2023vital}.} Test result on PURE~\cite{stricker2014non}, UBFC-rPPG~\cite{bobbia2019unsupervised} and MMPD-Simple~\cite{tang2023mmpd} trained on VV100~\cite{toye2023vital} and three subsets of VV100~\cite{toye2023vital} using TS-CAN~\cite{liu2020multi}, PhysNet~\cite{yu2019remote}, and PhysFormer~\cite{yu2022physformer}.}
	\vspace{0.1cm}
	\label{table2:training on vv100}
	\centering
	\small
	\setlength\tabcolsep{3pt} % default value: 6pt
	\begin{tabular}{r|cccccccccccc}
	\toprule
	  \textbf{Training Set} & \multicolumn{12}{c}{\textbf{VV100~\cite{toye2023vital}}} \\
	\textbf{Testing Set} & \multicolumn{12}{c}{\textbf{PURE~\cite{stricker2014non}}}  \\
        \textbf{Method} & \multicolumn{4}{c|}{TS-CAN~\cite{liu2020multi}} &\multicolumn{4}{c|}{PhysNet~\cite{yu2019remote}} &\multicolumn{4}{c}{PhysFormer~\cite{yu2022physformer}} \\
        \textbf{Metric} & MAE & RMSE & MAPE & \multicolumn{1}{c|}{$\rho$} & MAE & RMSE & MAPE & \multicolumn{1}{c|}{$\rho$} & MAE & RMSE & MAPE & $\rho$  \\ \hline  \hline 
         \textbf{Training Skin Tone} \\
         1 2(num:23)         & 6.22 & 17.25 & 7.41  & \multicolumn{1}{c|}{0.69} & \textbf{3.98}  & \textbf{13.96} & \textbf{3.44}  & \multicolumn{1}{c|}{\textbf{0.81}} & 4.74  & 16.01 & 4.67  & 0.74 \\
        3 4(num:23)         & 8.30 & 21.22 & 9.05  & \multicolumn{1}{c|}{0.48} & 5.60  & 15.35 & 7.58  & \multicolumn{1}{c|}{0.75} & 6.00  & 15.94 & 7.26  & 0.73 \\
        5 6(num:23)         & 9.21 & 21.03 & 10.84 & \multicolumn{1}{c|}{0.51} & 7.23  & 19.72 & 6.34  & \multicolumn{1}{c|}{0.60} & 15.06 & 24.06 & 21.83 & 0.17 \\
         1 2 3 4 5 6(num:100) & 6.40  & 15.99 & 8.87 & \multicolumn{1}{c|}{0.74} & 6.04  & 17.86 & 5.78  & \multicolumn{1}{c|}{0.66} & 5.47 & 15.84 & 6.11 & 0.74\\
        \bottomrule
        \end{tabular}\\
        \vspace{5pt}
        \begin{tabular}{r|cccccccccccc}
	\toprule
	  \textbf{Training Set} & \multicolumn{12}{c}{\textbf{VV100~\cite{toye2023vital}}} \\
	\textbf{Testing Set} & \multicolumn{12}{c}{\textbf{UBFC-rPPG}~\cite{bobbia2019unsupervised}}  \\
        \textbf{Method} & \multicolumn{4}{c|}{TS-CAN~\cite{liu2020multi}} &\multicolumn{4}{c|}{PhysNet~\cite{yu2019remote}} &\multicolumn{4}{c}{PhysFormer~\cite{yu2022physformer}} \\
        \textbf{Metric} & MAE & RMSE & MAPE & \multicolumn{1}{c|}{$\rho$} & MAE & RMSE & MAPE & \multicolumn{1}{c|}{$\rho$} & MAE & RMSE & MAPE & $\rho$\\ \hline  \hline 
         \textbf{Training Skin Tone} \\
         1 2(num:23)         & 3.58  & 10.72 & 3.37  & \multicolumn{1}{c|}{0.83}  & 5.63  & 14.64 & 5.01  & \multicolumn{1}{c|}{0.68}  & 6.36  & 15.64 & 5.72  & 0.67 \\
        3 4(num:23)         & 4.62  & 11.95 & 4.42  & \multicolumn{1}{c|}{0.80} & 2.03  & 5.91  & 2.05  & \multicolumn{1}{c|}{0.95} & 5.34  & 13.36 & 4.98  & 0.78 \\
        5 6(num:23)         & 11.26 & 21.14 & 10.64 & \multicolumn{1}{c|}{0.57} & 7.49  & 16.25 & 6.57  & \multicolumn{1}{c|}{0.61} & 20.84 & 27.89 & 19.50 & 0.23 \\
        1 2 3 4 5 6(num:100) & \textbf{1.70}  & \textbf{3.69}  & \textbf{1.83}  & \multicolumn{1}{c|}{\textbf{0.98}}  & 7.62  & 18.78 & 6.52  & \multicolumn{1}{c|}{0.48} & 2.64  & 7.05  & 2.49  & 0.93\\
        \bottomrule
        \end{tabular}
        \vspace{5pt}

        \begin{tabular}{r|cccccccccccc}
	\toprule
	  \textbf{Training Set} & \multicolumn{12}{c}{\textbf{VV100~\cite{toye2023vital}}} \\
	\textbf{Testing Set} & \multicolumn{12}{c}{\textbf{MMPD-Simple}~\cite{tang2023mmpd}}  \\
        \textbf{Method} & \multicolumn{4}{c|}{TS-CAN~\cite{liu2020multi}} &\multicolumn{4}{c|}{PhysNet~\cite{yu2019remote}} &\multicolumn{4}{c}{PhysFormer~\cite{yu2022physformer}} \\
        \textbf{Metric} & MAE & RMSE & MAPE & \multicolumn{1}{c|}{$\rho$} & MAE & RMSE & MAPE & \multicolumn{1}{c|}{$\rho$} & MAE & RMSE & MAPE & $\rho$\\ \hline  \hline 
         \textbf{Training Skin Tone} \\
         1 2(num:23)         & 4.23 & 8.59  & 5.51 & \multicolumn{1}{c|}{0.73} & 5.14 & 10.10 & 6.60 & \multicolumn{1}{c|}{0.58} & 6.90  & 12.36 & 8.54  & 0.37  \\
        3 4(num:23)         & 5.67 & 11.98 & 7.54 & \multicolumn{1}{c|}{0.42} & 5.49 & 10.83 & 8.51 & \multicolumn{1}{c|}{0.53} & 10.59 & 16.91 & 15.43 & -0.02 \\
        5 6(num:23)         & \textbf{4.07} & \textbf{9.36}  & \textbf{5.50} & \multicolumn{1}{c|}{\textbf{0.68}} & 4.62 & 10.33 & 5.44 & \multicolumn{1}{c|}{0.59} & 12.52 & 16.66 & 16.83 & 0.12  \\
        1 2 3 4 5 6(num:100) & 4.73 & 11.96 & 6.48  & \multicolumn{1}{c|}{0.48} & 4.70  & 9.93  & 5.87  & \multicolumn{1}{c|}{0.62} & 5.49 & 11.58 & 6.96 & 0.44\\
        \bottomrule
        \end{tabular}\\
       \footnotesize
       MAE = Mean Absolute Error in HR estimation (Beats/Min), RSME = Root Mean Square Error, MAPE = Mean Percentage Error (\%), $\rho$ = Pearson Correlation in HR estimation.
    \vspace{-0.4cm}
\end{table*}
\vspace{-0.1cm}

To investigate the variances in supervised neural network training across different skin tones, we divided the VV100~\cite{toye2023vital} dataset into three segments, each containing 23 subjects. Given the uneven distribution of skin tones, we created subsets pairing type 1 with type 2 skin tones, type 3 with type 4, and type 5 with type 6. Subsequently, we utilized three supervised learning methods to train on the entire VV100~\cite{toye2023vital} dataset and its three subsets individually. Table \ref{table2:training on vv100} presents the outcomes obtained by training on VV100~\cite{toye2023vital} and testing on three presentative datasets: PURE~\cite{stricker2014non}, UBFC-rPPG~\cite{bobbia2019unsupervised}, and MMPD-Simple~\cite{tang2023mmpd}. The best MAE indexes for PURE~\cite{stricker2014non}, UBFC-rPPG~\cite{bobbia2019unsupervised}, and MMPD-Simple~\cite{tang2023mmpd} are 3.98, 1.70, and 4.07 beats per minute in remote heart rate estimation task.

% \begin{figure}[htp]
%     \centering
%     \includegraphics[width=1\linewidth]{VV100result.jpg}
%     \caption{\textbf{Test Results Training on VV100.} Three notable test results are shown in Bland-Altman plots and their example curves on PURE, UBFC-rPPG and MMPD-Simple.}
%     \label{fig:VV100}
% \end{figure}

Test results on PURE are more accurate ~\cite{stricker2014non} when models are trained on lighter skin tones, outperforming those trained on the entire VV100~\cite{toye2023vital} dataset. For UBFC-rPPG~\cite{bobbia2019unsupervised}, which consists of type 1 to 3 skin tones, models trained on similar skin tones significantly outperformed those trained on darker skin tones. Notably, models trained on type 5 and type 6 skin tones showcased remarkably best test results on MMPD~\cite{tang2023mmpd}, especially with TS-CAN~\cite{liu2020multi} and PhysNet~\cite{yu2019remote}.  This underscores the pivotal role of maintaining consistency in skin color between the training and test sets, as it can significantly enhance test accuracy. Furthermore, skin tone diversity appears to contribute to more stable performance, achieving the best test results with TS-CAN~\cite{liu2020multi} on UBFC-rPPG~\cite{bobbia2019unsupervised}.
% The ground truth and prediction for the best results training on VV100 are illustrated in Fig \ref{fig:VV100}.
\vspace{-0.2cm}
\subsection{Supervised methods test on VV100~\cite{toye2023vital}} %Train ：MMPD（Simplest Subset）,PURE,UBFC-rPPG：3
% Test： vv100 all and single 6 skin tones：7
%TSCAN（2D+TSM） Physnet（3D） Spike-PhysFormer（Transformer）：3
\vspace{-0.1cm}
The results presented in Table \ref{table3: testing on vv100} on VV100~\cite{toye2023vital} delineate the performance of models trained across three datasets, with evaluations conducted on both aggregated and individual skin tone categories within the VV100~\cite{toye2023vital} dataset. Clearly, all models showed notable effectiveness on lighter skin tones, particularly TS-CAN~\cite{liu2020multi} and PhysNet~\cite{yu2019remote}. However, a discernible degradation in performance was observed as the skin tone deepened. Notably, the models trained on the MMPD-Simple~\cite{tang2023mmpd} dataset exhibited no superior performance on darker skin tones compared to its counterparts likely due to the compression of mobile phone shots. 
\vspace{-0.2cm}
\begin{table*}[h!]
    \small
	\caption{\textbf{Supervised methods test on VV100~\cite{toye2023vital}.} MAE on each single skin tone of VV100~\cite{toye2023vital} and entire VV100~\cite{toye2023vital} with models trained on PURE~\cite{stricker2014non}, UBFC-rPPG~\cite{bobbia2019unsupervised}, and MMPD-Simple~\cite{tang2023mmpd}.}
	\vspace{0.1cm}
	\label{table3: testing on vv100}
	\centering
	\small
	\setlength\tabcolsep{3pt} % default value: 6pt
        \begin{tabular}{r|ccccccccc}
	\toprule
	  \textbf{Testing Set} & \multicolumn{9}{c}{\textbf{VV100}~\cite{toye2023vital}}  \\
   \textbf{Training Set} & \multicolumn{3}{c|}{\textbf{PURE~\cite{stricker2014non}}} & \multicolumn{3}{c|}{\textbf{UBFC-rPPG}~\cite{bobbia2019unsupervised}} & \multicolumn{3}{c}{\textbf{MMPD-Simple}~\cite{tang2023mmpd}}\\
	
        \textbf{Method} & TC~\cite{liu2020multi} &PN~\cite{yu2019remote} &\multicolumn{1}{c|}{PF~\cite{yu2022physformer}} & TC~\cite{liu2020multi} &PN~\cite{yu2019remote} &\multicolumn{1}{c|}{PF~\cite{yu2022physformer}} & TC~\cite{liu2020multi} &PN~\cite{yu2019remote} &PF~\cite{yu2022physformer} \\
        \hline  \hline 
         \textbf{Testing Skin Tone} \\
         1(num:5)           & 0.22 &\textbf{0.00} &  \multicolumn{1}{c|}{0.59} & 0.66 & 0.66&  \multicolumn{1}{c|}{\textbf{0.59}}  & \textbf{0.00} & 0.00 & 0.59  \\
        2(num:35)           & 3.24  & \textbf{1.58} &  \multicolumn{1}{c|}{2.26}  &\textbf{1.39}  & 5.50 &  \multicolumn{1}{c|}{2.26} & \textbf{2.67} & 3.02 & 4.10 \\
        3(num:21)           & 9.24 & \textbf{1.35} &  \multicolumn{1}{c|}{7.57}  & \textbf{4.33} & 5.68 &  \multicolumn{1}{c|}{7.57} & \textbf{6.70}  & 7.32 & 6.90 \\
        4(num:17)           & 6.15 & \textbf{4.34}   & \multicolumn{1}{c|}{10.62} & 5.49 & \textbf{3.46}   &  \multicolumn{1}{c|}{10.62} & 7.03 & \textbf{6.26}  & 7.69 \\
        5(num:13)           & 13.00& \textbf{13.02} & \multicolumn{1}{c|}{11.91} & 10.11  & \textbf{9.35} &  \multicolumn{1}{c|}{11.91}  & 10.23  & \textbf{8.60} & 14.94 \\
        6(num:9)           & \textbf{14.26}  & 14.65 & \multicolumn{1}{c|}{17.23} & 12.40 & \textbf{11.33}  &  \multicolumn{1}{c|}{17.23}& 15.72  & 16.11 & \textbf{15.12}\\
        1 2 3 4 5 6(num:100) & 7.23& \textbf{3.78}  & \multicolumn{1}{c|}{7.29} & \textbf{4.86} & 6.08  &  \multicolumn{1}{c|}{7.83}   & 6.36 & \textbf{6.30} & 7.58  \\
        \bottomrule
        \end{tabular}\\
        \vspace{5pt}
       \footnotesize
       TC, PN, and PF stand for  TS-CAN~\cite{liu2020multi}, PhysNet~\cite{yu2019remote}, and PhysFormer~\cite{yu2022physformer}. MAE = Mean Absolute Error in HR estimation (Beats/Min)
    \vspace{-0.3cm}
\end{table*}

\vspace{-0.5cm}
\section{Discussion}
\vspace{-0.1cm}
\subsection{Supervised methods on VVALL~\cite{toye2023vital} and UBFC-rPPG~\cite{bobbia2019unsupervised}, PURE~\cite{stricker2014non}, and MMPD-Simple~\cite{tang2023mmpd} for data volume}
\begin{figure}[htbp]
    \centering
    \begin{subfigure}{0.5\textwidth}
        \centering
        \includegraphics[width=\linewidth]{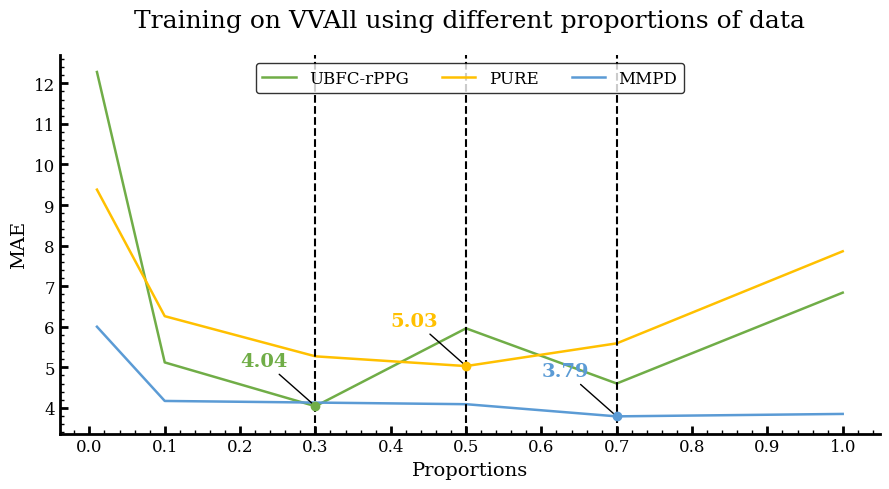}
        \caption{\textbf{Training on VVAll~\cite{toye2023vital} using different proportions of data.} Comparison of test results on PURE~\cite{stricker2014non}, UBFC-rPPG~\cite{bobbia2019unsupervised}, and MMPD-Simple~\cite{tang2023mmpd} from training on different proportions of VVAll~\cite{toye2023vital}.}
        \label{fig:VVAll}
    \end{subfigure}%
    \begin{subfigure}{0.5\textwidth}
        \centering
        \includegraphics[width=\linewidth]{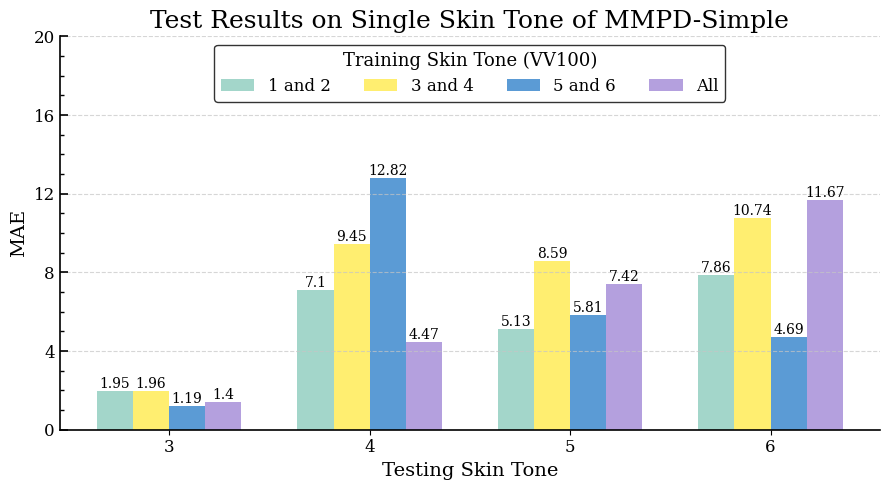}
        \caption{\textbf{Training on VV100~\cite{toye2023vital} and testing on MMPD-Simple~\cite{tang2023mmpd} by single skin tone.} Comparison of test results on each single skin tone on MMPD-Simple~\cite{tang2023mmpd} with models trained on three subsets of VV100~\cite{toye2023vital}.}
        \label{fig:testMMPD}
    \end{subfigure}
    \caption{Training on VV100~\cite{toye2023vital} and VVAll~\cite{toye2023vital} with TS-CAN~\cite{liu2020multi}. MAE = Mean Absolute Error in HR estimation (Beats/Min)}
    \label{fig:Discussion}
  \vspace{-0.8cm}  
\end{figure}

\vspace{-0.1cm}
%train: vvall 1%,10%,100%
%test: UBFC,PURE
%TSCAN（2D+TSM） Physnet（3D） Spike-PhysFormer（Transformer）：3
We investigated the impact of different proportions of training data in the VVAll~\cite{toye2023vital} dataset analysis. Specifically, we trained models on datasets comprising 100\%, 70\%, 50\%, 30\%, 10\%, and 1\% of the VVAll~\cite{toye2023vital} dataset, corresponding to 893, 625, 446, 267, 89, and 8 subjects respectively, subsequently tested them on UBFC-rPPG~\cite{bobbia2019unsupervised}, PURE~\cite{stricker2014non}, and MMPD-Simple~\cite{tang2023mmpd}, using TS-CAN~\cite{liu2020multi}. Notably, for the 10\% and 1\% proportions, we divided the dataset into five distinct parts and averaged their outcomes to minimize the impact of random selection. It is evident from Fig \ref{fig:VVAll} that an increased volume of training data does not necessarily lead to improved model performance. Instead, a threshold exists beyond which adding more subjects to the training set ceases to significantly improve model performance. Importantly, this threshold varies with the complexity of the testing dataset but remains below 1000.
\vspace{-0.3cm}

\vspace{-0.3cm}
\subsection{Supervised methods on VV100~\cite{toye2023vital} and MMPD-Simple~\cite{tang2023mmpd} for diversity}
\vspace{-0.1cm}

In addition to quantity, the dimension of diversity also merits attention. Specifically, we conducted tests on various skin tones within the MMPD-Simple~\cite{tang2023mmpd} dataset in isolation, utilizing a model trained on the VV100~\cite{toye2023vital} dataset. The results are illustrated in Fig \ref{fig:testMMPD}. For lighter skin tones (i.e. types 3 and 4), diversity proves more crucial than skin tone consistency. Conversely, for darker skin tones (i.e., types 5 and 6), training on similar skin types often yields better results than training with a larger number of subjects.

% \begin{table*}[h!]
%     \small
% 	\caption{\textbf{Training on VV100~\cite{toye2023vital}.} }
% 	\vspace{0.3cm}
% 	\label{table5:single skin tone}
% 	\centering
% 	\small
% 	\setlength\tabcolsep{3pt} % default value: 6pt
%         \begin{tabular}{r|cccccccc}
% 	\toprule
%         \textbf{Training Set} & \multicolumn{8}{c}{\textbf{VV100~\cite{toye2023vital}}} \\
% 	  \textbf{Method} & \multicolumn{8}{c}{\textbf{TS-CAN~\cite{liu2020multi}}}  \\
%         \textbf{Testing Set} &\multicolumn{8}{c}{MMPD-Simple~\cite{tang2023mmpd}}\\
%         \textbf{Testing Skin Tone} & \multicolumn{2}{c|}{3} &\multicolumn{2}{c|}{4} &\multicolumn{2}{c|}{5}&\multicolumn{2}{c}{6} \\
%         \textbf{Metric} & MAE & \multicolumn{1}{c|}{$\rho$} & MAE & \multicolumn{1}{c|}{$\rho$}& MAE & \multicolumn{1}{c|}{$\rho$} & MAE  & $\rho$\\ \hline  \hline 
%         \textbf{Training Skin Tone}  \\
%         1 2(num:23)         & 1.95 & \multicolumn{1}{c|}{0.92} & 7.10  & \multicolumn{1}{c|}{0.54} & 5.13  & \multicolumn{1}{c|}{0.79} & 7.86  & 0.51\\
%         3 4(num:23)        & 1.96  & \multicolumn{1}{c|}{0.92} & 9.45   & \multicolumn{1}{c|}{0.23} & 8.59  & \multicolumn{1}{c|}{0.12} & 10.74 & 0.23\\
%         5 6(num:23)        & 1.19 & \multicolumn{1}{c|}{0.96} & 12.82  & \multicolumn{1}{c|}{0.20}  & 5.81 & \multicolumn{1}{c|}{0.52} & 4.69 &0.73\\
%         1 2 3 4 5 6 (num:100)& 1.40 & \multicolumn{1}{c|}{0.97} & 4.47  & \multicolumn{1}{c|}{0.65} & 7.42 & \multicolumn{1}{c|}{0.10} & 11.67 & 0.13\\
%         \bottomrule
%         \end{tabular}\\
%         \footnotesize
%     \vspace{-0.3cm}
% \end{table*}
\vspace{-0.3cm}
\subsection{Limitation}
\vspace{-0.1cm}
We recognize the presence of certain limitations within our study and are committed to addressing these in future research efforts. The primary limitations identified are as follows: 1) New neural network modifications necessary for achieving state-of-the-art performance were not implemented. 2) There is an absence of a detailed exploration into the optical and medical principles that could explain the models' underperformance in challenging scenarios. 

\vspace{-0.5cm}
\section{Conclusion}
\vspace{-0.2cm}
This study marks the inaugural exploration of the VitalVideo dataset~\cite{toye2023vital}, the most extensive rPPG dataset to date. Leveraging its vast array of subjects and diverse skin tones, we embarked on a comprehensive series of experiments employing six unsupervised methods alongside three supervised models. Our key findings include: 1) Our pioneering training and testing on the VitalVideo dataset, coupled with the open sharing of our code, underscores the dataset's reliability for advancing research in this field.
2) Optimal rPPG model training is achievable with datasets comprising just a few hundred individuals, demonstrating the efficiency of smaller, well-curated datasets.
3) We discovered a pronounced sensitivity to skin tone consistency, particularly with darker skin tones, highlighting the critical need for careful consideration of skin tone diversity in model development and evaluation.

\vspace{-0.5cm}
\section{Acknowledgments}
This work is supported by National Natural Science Foundation of China (No. 62132010 and No. 62002198), Young Elite Scientists Sponsorship Program by CAST (No. 2021QNRC001), and Beijing Natural Science Foundation (No. QY23124).

%%
%% Define the bibliography file to be used
\bibliography{sample-ceur}

\end{document}